\documentclass[letterpaper, 10 pt, conference]{ieeeconf}  
\IEEEoverridecommandlockouts    
\usepackage{times}
\usepackage{graphicx} 
\usepackage{algorithm}
\usepackage{algorithmic}
\usepackage{hyperref}

\usepackage{url}            
\usepackage{nicefrac}       
\usepackage{amssymb}
\usepackage{amsmath}
\usepackage{paralist}
\usepackage{subcaption}
\usepackage{setspace}
\usepackage{booktabs}
\usepackage{microtype}
\usepackage{multirow}
\usepackage{array}
\usepackage{balance}

\usepackage{amssymb}
\usepackage{amsfonts}
\usepackage{mathrsfs}
\usepackage{xspace}
\usepackage{bm}
\usepackage{upgreek}

\newcommand{\safemath}[2]{\newcommand{#1}{\ensuremath{#2}\xspace}}

\safemath{\bma}{\mathbf{a}}
\safemath{\bmb}{\mathbf{b}}
\safemath{\bmc}{\mathbf{c}}
\safemath{\bmd}{\mathbf{d}}
\safemath{\bme}{\mathbf{e}}
\safemath{\bmf}{\mathbf{f}}
\safemath{\bmg}{\mathbf{g}}
\safemath{\bmh}{\mathbf{h}}
\safemath{\bmi}{\mathbf{i}}
\safemath{\bmj}{\mathbf{j}}
\safemath{\bmk}{\mathbf{k}}
\safemath{\bml}{\mathbf{l}}
\safemath{\bmm}{\mathbf{m}}
\safemath{\bmn}{\mathbf{n}}
\safemath{\bmo}{\mathbf{o}}
\safemath{\bmp}{\mathbf{p}}
\safemath{\bmq}{\mathbf{q}}
\safemath{\bmr}{\mathbf{r}}
\safemath{\bms}{\mathbf{s}}
\safemath{\bmt}{\mathbf{t}}
\safemath{\bmu}{\mathbf{u}}
\safemath{\bmv}{\mathbf{v}}
\safemath{\bmw}{\mathbf{w}}
\safemath{\bmx}{\mathbf{x}}
\safemath{\bmy}{\mathbf{y}}
\safemath{\bmz}{\mathbf{z}}
\safemath{\bmzero}{\mathbf{0}}
\safemath{\bmone}{\mathbf{1}}

\bmdefine{\biad}{a}
\bmdefine{\bibd}{b}
\bmdefine{\bicd}{c}
\bmdefine{\bidd}{d}
\bmdefine{\bied}{e}
\bmdefine{\bifd}{f}
\bmdefine{\bigd}{g}
\bmdefine{\bihd}{h}
\bmdefine{\biid}{i}
\bmdefine{\bijd}{j}
\bmdefine{\bikd}{k}
\bmdefine{\bild}{l}
\bmdefine{\bimd}{m}
\bmdefine{\bind}{n}
\bmdefine{\biod}{o}
\bmdefine{\bipd}{p}
\bmdefine{\biqd}{q}
\bmdefine{\bird}{r}
\bmdefine{\bisd}{s}
\bmdefine{\bitd}{t}
\bmdefine{\biud}{u}
\bmdefine{\bivd}{v}
\bmdefine{\biwd}{w}
\bmdefine{\bixd}{x}
\bmdefine{\biyd}{y}
\bmdefine{\bizd}{z}

\bmdefine{\bixid}{\xi}
\bmdefine{\bilambdad}{\lambda}
\bmdefine{\bimud}{\mu}
\bmdefine{\bithetad}{\theta}
\bmdefine{\biphid}{\phi}
\bmdefine{\bideltad}{\delta}

\safemath{\bmia}{\biad}
\safemath{\bmib}{\bibd}
\safemath{\bmic}{\bicd}
\safemath{\bmid}{\bidd}
\safemath{\bmie}{\bied}
\safemath{\bmif}{\bifd}
\safemath{\bmig}{\bigd}
\safemath{\bmih}{\bihd}
\safemath{\bmii}{\biid}
\safemath{\bmij}{\bijd}
\safemath{\bmik}{\bikd}
\safemath{\bmil}{\bild}
\safemath{\bmim}{\bimd}
\safemath{\bmin}{\bind}
\safemath{\bmio}{\biod}
\safemath{\bmip}{\bipd}
\safemath{\bmiq}{\biqd}
\safemath{\bmir}{\bird}
\safemath{\bmis}{\bisd}
\safemath{\bmit}{\bitd}
\safemath{\bmiu}{\biud}
\safemath{\bmiv}{\bivd}
\safemath{\bmiw}{\biwd}
\safemath{\bmix}{\bixd}
\safemath{\bmiy}{\biyd}
\safemath{\bmiz}{\bizd}

\safemath{\bmxi}{\bixid}
\safemath{\bmlambda}{\bilambdad}
\safemath{\bmmu}{\bimud}
\safemath{\bmtheta}{\bithetad}
\safemath{\bmphi}{\biphid}
\safemath{\bmdelta}{\bideltad}

\safemath{\bA}{\mathbf{A}}
\safemath{\bB}{\mathbf{B}}
\safemath{\bC}{\mathbf{C}}
\safemath{\bD}{\mathbf{D}}
\safemath{\bE}{\mathbf{E}}
\safemath{\bF}{\mathbf{F}}
\safemath{\bG}{\mathbf{G}}
\safemath{\bH}{\mathbf{H}}
\safemath{\bI}{\mathbf{I}}
\safemath{\bJ}{\mathbf{J}}
\safemath{\bK}{\mathbf{K}}
\safemath{\bL}{\mathbf{L}}
\safemath{\bM}{\mathbf{M}}
\safemath{\bN}{\mathbf{N}}
\safemath{\bO}{\mathbf{O}}
\safemath{\bP}{\mathbf{P}}
\safemath{\bQ}{\mathbf{Q}}
\safemath{\bR}{\mathbf{R}}
\safemath{\bS}{\mathbf{S}}
\safemath{\bT}{\mathbf{T}}
\safemath{\bU}{\mathbf{U}}
\safemath{\bV}{\mathbf{V}}
\safemath{\bW}{\mathbf{W}}
\safemath{\bX}{\mathbf{X}}
\safemath{\bY}{\mathbf{Y}}
\safemath{\bZ}{\mathbf{Z}}

\safemath{\bZero}{\mathbf{0}}
\safemath{\bOne}{\mathbf{1}}
\safemath{\bDelta}{\mathbf{\Delta}}
\safemath{\bLambda}{\boldsymbol\Lambda}
\safemath{\bPhi}{\mathbf{\Upphi}}
\safemath{\bSigma}{\mathbf{\Upsigma}}
\safemath{\bOmega}{\mathbf{\Upomega}}
\safemath{\bTheta}{\mathbf{\Uptheta}}

\bmdefine{\biAd}{A}
\bmdefine{\biBd}{B}
\bmdefine{\biCd}{C}
\bmdefine{\biDd}{D}
\bmdefine{\biEd}{E}
\bmdefine{\biFd}{F}
\bmdefine{\biGd}{G}
\bmdefine{\biHd}{H}
\bmdefine{\biId}{I}
\bmdefine{\biJd}{J}
\bmdefine{\biKd}{K}
\bmdefine{\biLd}{L}
\bmdefine{\biMd}{M}
\bmdefine{\biOd}{N}
\bmdefine{\biPd}{O}
\bmdefine{\biQd}{P}
\bmdefine{\biRd}{R}
\bmdefine{\biSd}{S}
\bmdefine{\biTd}{T}
\bmdefine{\biUd}{U}
\bmdefine{\biVd}{V}
\bmdefine{\biWd}{W}
\bmdefine{\biXd}{X}
\bmdefine{\biYd}{Y}
\bmdefine{\biZd}{Z}

\bmdefine{\biDelta}{\Delta}
\bmdefine{\biLambda}{\Lambda}
\bmdefine{\biPhi}{\Phi}
\bmdefine{\biSigma}{\Sigma}
\bmdefine{\biOmega}{\Omega}
\bmdefine{\biTheta}{\Theta}

\safemath{\bimA}{\biAd}
\safemath{\bimB}{\biBd}
\safemath{\bimC}{\biCd}
\safemath{\bimD}{\biDd}
\safemath{\bimE}{\biEd}
\safemath{\bimF}{\biFd}
\safemath{\bimG}{\biGd}
\safemath{\bimH}{\biHd}
\safemath{\bimI}{\biId}
\safemath{\bimJ}{\biJd}
\safemath{\bimK}{\biKd}
\safemath{\bimL}{\biLd}
\safemath{\bimM}{\biMd}
\safemath{\bimN}{\biNd}
\safemath{\bimO}{\biOd}
\safemath{\bimP}{\biPd}
\safemath{\bimQ}{\biQd}
\safemath{\bimR}{\biRd}
\safemath{\bimS}{\biSd}
\safemath{\bimT}{\biTd}
\safemath{\bimU}{\biUd}
\safemath{\bimV}{\biVd}
\safemath{\bimW}{\biWd}
\safemath{\bimX}{\biXd}
\safemath{\bimY}{\biYd}
\safemath{\bimZ}{\biZd}

\safemath{\bimDelta}{\biDelta}
\safemath{\bimLambda}{\biLambda}
\safemath{\bimPhi}{\biPhi}
\safemath{\bimSigma}{\biSigma}
\safemath{\bimOmega}{\biOmega}
\safemath{\bimTheta}{\biTheta}

\safemath{\setA}{\mathcal{A}}
\safemath{\setB}{\mathcal{B}}
\safemath{\setC}{\mathcal{C}}
\safemath{\setD}{\mathcal{D}}
\safemath{\setE}{\mathcal{E}}
\safemath{\setF}{\mathcal{F}}
\safemath{\setG}{\mathcal{G}}
\safemath{\setH}{\mathcal{H}}
\safemath{\setI}{\mathcal{I}}
\safemath{\setJ}{\mathcal{J}}
\safemath{\setK}{\mathcal{K}}
\safemath{\setL}{\mathcal{L}}
\safemath{\setM}{\mathcal{M}}
\safemath{\setN}{\mathcal{N}}
\safemath{\setO}{\mathcal{O}}
\safemath{\setP}{\mathcal{P}}
\safemath{\setQ}{\mathcal{Q}}
\safemath{\setR}{\mathcal{R}}
\safemath{\setS}{\mathcal{S}}
\safemath{\setT}{\mathcal{T}}
\safemath{\setU}{\mathcal{U}}
\safemath{\setV}{\mathcal{V}}
\safemath{\setW}{\mathcal{W}}
\safemath{\setX}{\mathcal{X}}
\safemath{\setY}{\mathcal{Y}}
\safemath{\setZ}{\mathcal{Z}}
\safemath{\emptySet}{\varnothing}

\safemath{\colA}{\mathscr{A}}
\safemath{\colB}{\mathscr{B}}
\safemath{\colC}{\mathscr{C}}
\safemath{\colD}{\mathscr{D}}
\safemath{\colE}{\mathscr{E}}
\safemath{\colF}{\mathscr{F}}
\safemath{\colG}{\mathscr{G}}
\safemath{\colH}{\mathscr{H}}
\safemath{\colI}{\mathscr{I}}
\safemath{\colJ}{\mathscr{J}}
\safemath{\colK}{\mathscr{K}}
\safemath{\colL}{\mathscr{L}}
\safemath{\colM}{\mathscr{M}}
\safemath{\colN}{\mathscr{N}}
\safemath{\colO}{\mathscr{O}}
\safemath{\colP}{\mathscr{P}}
\safemath{\colQ}{\mathscr{Q}}
\safemath{\colR}{\mathscr{R}}
\safemath{\colS}{\mathscr{S}}
\safemath{\colT}{\mathscr{T}}
\safemath{\colU}{\mathscr{U}}
\safemath{\colV}{\mathscr{V}}
\safemath{\colW}{\mathscr{W}}
\safemath{\colX}{\mathscr{X}}
\safemath{\colY}{\mathscr{Y}}
\safemath{\colZ}{\mathscr{Z}}

\safemath{\opA}{\mathbb{A}}
\safemath{\opB}{\mathbb{B}}
\safemath{\opC}{\mathbb{C}}
\safemath{\opD}{\mathbb{D}}
\safemath{\opE}{\mathbb{E}}
\safemath{\opF}{\mathbb{F}}
\safemath{\opG}{\mathbb{G}}
\safemath{\opH}{\mathbb{H}}
\safemath{\opI}{\mathbb{I}}
\safemath{\opJ}{\mathbb{J}}
\safemath{\opK}{\mathbb{K}}
\safemath{\opL}{\mathbb{L}}
\safemath{\opM}{\mathbb{M}}
\safemath{\opN}{\mathbb{N}}
\safemath{\opO}{\mathbb{O}}
\safemath{\opP}{\mathbb{P}}
\safemath{\opQ}{\mathbb{Q}}
\safemath{\opR}{\mathbb{R}}
\safemath{\opS}{\mathbb{S}}
\safemath{\opT}{\mathbb{T}}
\safemath{\opU}{\mathbb{U}}
\safemath{\opV}{\mathbb{V}}
\safemath{\opW}{\mathbb{W}}
\safemath{\opX}{\mathbb{X}}
\safemath{\opY}{\mathbb{Y}}
\safemath{\opZ}{\mathbb{Z}}
\safemath{\opZero}{\mathbb{O}}
\safemath{\identityop}{\opI}

\safemath{\veca}{\bma}
\safemath{\vecb}{\bmb}
\safemath{\vecc}{\bmc}
\safemath{\vecd}{\bmd}
\safemath{\vece}{\bme}
\safemath{\vecf}{\bmf}
\safemath{\vecg}{\bmg}
\safemath{\vech}{\bmh}
\safemath{\veci}{\bmi}
\safemath{\vecj}{\bmj}
\safemath{\veck}{\bmk}
\safemath{\vecl}{\bml}
\safemath{\vecm}{\bmm}
\safemath{\vecn}{\bmn}
\safemath{\veco}{\bmo}
\safemath{\vecp}{\bmp}
\safemath{\vecq}{\bmq}
\safemath{\vecr}{\bmr}
\safemath{\vecs}{\bms}
\safemath{\vect}{\bmt}
\safemath{\vecu}{\bmu}
\safemath{\vecv}{\bmv}
\safemath{\vecw}{\bmw}
\safemath{\vecx}{\bmx}
\safemath{\vecy}{\bmy}
\safemath{\vecz}{\bmz}

\safemath{\veczero}{\bmzero}
\safemath{\vecone}{\bmone}
\safemath{\vecxi}{\bmxi}
\safemath{\veclambda}{\bmlambda}
\safemath{\vecmu}{\bmmu}
\safemath{\vectheta}{\bmtheta}
\safemath{\vecphi}{\bmphi}
\safemath{\vecdelta}{\bmdelta}

\safemath{\matA}{\bA}
\safemath{\matB}{\bB}
\safemath{\matC}{\bC}
\safemath{\matD}{\bD}
\safemath{\matE}{\bE}
\safemath{\matF}{\bF}
\safemath{\matG}{\bG}
\safemath{\matH}{\bH}
\safemath{\matI}{\bI}
\safemath{\matJ}{\bJ}
\safemath{\matK}{\bK}
\safemath{\matL}{\bL}
\safemath{\matM}{\bM}
\safemath{\matN}{\bN}
\safemath{\matO}{\bO}
\safemath{\matP}{\bP}
\safemath{\matQ}{\bQ}
\safemath{\matR}{\bR}
\safemath{\matS}{\bS}
\safemath{\matT}{\bT}
\safemath{\matU}{\bU}
\safemath{\matV}{\bV}
\safemath{\matW}{\bW}
\safemath{\matX}{\bX}
\safemath{\matY}{\bY}
\safemath{\matZ}{\bZ}
\safemath{\matzero}{\bmzero}

\safemath{\matDelta}{\bDelta}
\safemath{\matLambda}{\bLambda}
\safemath{\matPhi}{\bPhi}
\safemath{\matSigma}{\bSigma}
\safemath{\matOmega}{\bOmega}
\safemath{\matTheta}{\bTheta}

\safemath{\matidentity}{\matI}
\safemath{\matone}{\matO}

\safemath{\rnda}{A}
\safemath{\rndb}{B}
\safemath{\rndc}{C}
\safemath{\rndd}{D}
\safemath{\rnde}{E}
\safemath{\rndf}{F}
\safemath{\rndg}{G}
\safemath{\rndh}{H}
\safemath{\rndi}{I}
\safemath{\rndj}{J}
\safemath{\rndk}{K}
\safemath{\rndl}{L}
\safemath{\rndm}{M}
\safemath{\rndn}{N}
\safemath{\rndo}{O}
\safemath{\rndp}{P}
\safemath{\rndq}{Q}
\safemath{\rndr}{R}
\safemath{\rnds}{S}
\safemath{\rndt}{T}
\safemath{\rndu}{U}
\safemath{\rndv}{V}
\safemath{\rndw}{W}
\safemath{\rndx}{X}
\safemath{\rndy}{Y}
\safemath{\rndz}{Z}

\safemath{\rveca}{\bimA}
\safemath{\rvecb}{\bimB}
\safemath{\rvecc}{\bimC}
\safemath{\rvecd}{\bimD}
\safemath{\rvece}{\bimE}
\safemath{\rvecf}{\bimF}
\safemath{\rvecg}{\bimG}
\safemath{\rvech}{\bimH}
\safemath{\rveci}{\bimI}
\safemath{\rvecj}{\bimJ}
\safemath{\rveck}{\bimK}
\safemath{\rvecl}{\bimL}
\safemath{\rvecm}{\bimM}
\safemath{\rvecn}{\bimN}
\safemath{\rveco}{\bomO}
\safemath{\rvecp}{\bimP}
\safemath{\rvecq}{\bimQ}
\safemath{\rvecr}{\bimR}
\safemath{\rvecs}{\bimS}
\safemath{\rvect}{\bimT}
\safemath{\rvecu}{\bimU}
\safemath{\rvecv}{\bimV}
\safemath{\rvecw}{\bimW}
\safemath{\rvecx}{\bimX}
\safemath{\rvecy}{\bimY}
\safemath{\rvecz}{\bimZ}

\safemath{\rvecxi}{\bmxi}
\safemath{\rveclambda}{\bmlambda}
\safemath{\rvecmu}{\bmmu}
\safemath{\rvectheta}{\bmtheta}
\safemath{\rvecphi}{\bmphi}

\safemath{\rmatA}{\bimA}
\safemath{\rmatB}{\bimB}
\safemath{\rmatC}{\bimC}
\safemath{\rmatD}{\bimD}
\safemath{\rmatE}{\bimE}
\safemath{\rmatF}{\bimF}
\safemath{\rmatG}{\bimG}
\safemath{\rmatH}{\bimH}
\safemath{\rmatI}{\bimI}
\safemath{\rmatJ}{\bimJ}
\safemath{\rmatK}{\bimK}
\safemath{\rmatL}{\bimL}
\safemath{\rmatM}{\bimM}
\safemath{\rmatN}{\bimN}
\safemath{\rmatO}{\bimO}
\safemath{\rmatP}{\bimP}
\safemath{\rmatQ}{\bimQ}
\safemath{\rmatR}{\bimR}
\safemath{\rmatS}{\bimS}
\safemath{\rmatT}{\bimT}
\safemath{\rmatU}{\bimU}
\safemath{\rmatV}{\bimV}
\safemath{\rmatW}{\bimW}
\safemath{\rmatX}{\bimX}
\safemath{\rmatY}{\bimY}
\safemath{\rmatZ}{\bimZ}

\safemath{\rmatDelta}{\bimDelta}
\safemath{\rmatLambda}{\bimLambda}
\safemath{\rmatPhi}{\bimPhi}
\safemath{\rmatSigma}{\bimSigma}
\safemath{\rmatOmega}{\bimOmega}
\safemath{\rmatTheta}{\bimTheta}

\usepackage{amssymb}
\usepackage{amsfonts}
\usepackage{mathrsfs}
\usepackage{xspace}
\usepackage{bm}
\usepackage{fancyref}
\usepackage{textcomp}

\usepackage{multirow}
\usepackage{stmaryrd}

\newenvironment{textbmatrix}{	\setlength{\arraycolsep}{2.5pt}%
								\big[\begin{matrix}}{\end{matrix}\big]%
								\raisebox{0.08ex}{\vphantom{M}}}

\def\be{\begin{equation}}
\def\ee{\end{equation}}
\def\een{\nonumber \end{equation}}
\def\mat{\begin{bmatrix}}
\def\emat{\end{bmatrix}}
\def\btm{\begin{textbmatrix}}
\def\etm{\end{textbmatrix}}

\def\ba#1\ea{\begin{align}#1\end{align}}
\def\bas#1\eas{\begin{align*}#1\end{align*}}
\def\bs#1\es{\begin{split}#1\end{split}} 
\def\bg#1\eg{\begin{gather}#1\end{gather}}
\def\bml#1\eml{\begin{multline}#1\end{multline}}
\def\bi#1\ei{\begin{itemize}#1\end{itemize}}

\newcommand{\lefto}{\mathopen{}\left}

\DeclareMathOperator{\tr}{tr}				
\DeclareMathOperator{\diag}{diag}			
\DeclareMathOperator*{\argmax}{arg\;max}		
\DeclareMathOperator{\Exop}{\opE}			

\newcommand{\Ex}[2]{\ensuremath{\Exop_{#1}\lefto[#2\right]}} 	

\safemath{\dirac}{\delta}					
\safemath{\krond}{\dirac}					

\safemath{\upto}{\uparrow}
\safemath{\downto}{\downarrow}
\safemath{\iu}{j}							
\safemath{\ev}{\lambda}						
\safemath{\hilseqspace}{l^{2}}				
\newcommand{\banachfunspace}[1]{\setL^{#1}}	
\safemath{\hilfunspace}{\banachfunspace{2}}	

\safemath{\SNR}{\textsf{SNR}} 				
\safemath{\PAR}{\textsf{PAR}} 				
\safemath{\No}{N_0}							
\safemath{\Es}{E_s}							
\safemath{\Eb}{E_b}							
\safemath{\EbNo}{\frac{\Eb}{\No}}
\safemath{\EsNo}{\frac{\Es}{\No}}

\DeclareMathOperator{\CHop}{\ensuremath{\opH}} 
\safemath{\tvir}{\rndh_{\CHop}}				
\safemath{\tvtf}{\rndl_{\CHop}}				
\safemath{\spf}{\rnds_{\CHop}}				
\safemath{\bff}{H_{\CHop}}					

\safemath{\ircf}{r_{h}}						
\safemath{\tftvcf}{r_{s}}					
\safemath{\tfcf}{r_{l}}						
\safemath{\bfcf}{r_{H}}						

\safemath{\tcorr}{c_h}						
\safemath{\scf}{c_{s}}						
\safemath{\tfcorr}{c_{l}}					
\safemath{\fcorr}{c_{H}}						

\safemath{\mi}{I}							
\safemath{\capacity}{C}						

\safemath{\normal}{\mathcal{N}}			
\safemath{\jpg}{\mathcal{CN}}			
\safemath{\mchain}{\leftrightarrow}		

\safemath{\dB}{\,\mathrm{dB}}
\safemath{\dBm}{\,\mathrm{dBm}}
\safemath{\Hz}{\,\mathrm{Hz}}
\safemath{\kHz}{\,\mathrm{kHz}}
\safemath{\MHz}{\,\mathrm{MHz}}
\safemath{\GHz}{\,\mathrm{GHz}}
\safemath{\s}{\,\mathrm{s}}
\safemath{\ms}{\,\mathrm{ms}}
\safemath{\mus}{\,\mathrm{\text{\textmu}s}}
\safemath{\ns}{\,\mathrm{ns}}
\safemath{\ps}{\,\mathrm{ps}}
\safemath{\meter}{\,\mathrm{m}}
\safemath{\mm}{\,\mathrm{mm}}
\safemath{\cm}{\,\mathrm{cm}}
\safemath{\m}{\,\mathrm{m}}
\safemath{\W}{\,\mathrm{W}}
\safemath{\mW}{\, \mathrm{mW}}
\safemath{\J}{\,\mathrm{J}}
\safemath{\K}{\,\mathrm{K}}
\safemath{\bit}{\,\mathrm{bit}}
\safemath{\nat}{\,\mathrm{nat}}

\safemath{\define}{\triangleq}			

\safemath{\equivalent}{\sim}
\safemath{\distas}{\sim}					
\safemath{\sdiff}{\Delta}				

\safemath{\reals}{\mathbb{R}}
\safemath{\positivereals}{\reals_{+}}
\safemath{\integers}{\mathbb{Z}}
\safemath{\posint}{\integers_{+}}
\safemath{\naturals}{\mathbb{N}}
\safemath{\posnaturals}{\naturals_{+}}
\safemath{\complexset}{\mathbb{C}}
\safemath{\rationals}{\mathbb{Q}}

\newcommand*{\fancyrefapplabelprefix}{app}		
\newcommand*{\fancyrefthmlabelprefix}{thm}		
\newcommand*{\fancyreflemlabelprefix}{lem}		
\newcommand*{\fancyrefcorlabelprefix}{cor}		
\newcommand*{\fancyrefdeflabelprefix}{def}		
\newcommand*{\fancyrefproplabelprefix}{prop}	
\newcommand*{\fancyrefobslabelprefix}{obs}		
\newcommand*{\fancyrefalglabelprefix}{alg}		
\newcommand*{\fancyrefasmlabelprefix}{asm}	    
\newcommand*{\fancyrefasmslabelprefix}{asms}	    
\newcommand*{\fancyreftbllabelprefix}{tbl}	    
\newcommand*{\fancyrefestilabelprefix}{esti}	    

\frefformat{vario}{\fancyrefseclabelprefix}{Section~#1}
\frefformat{vario}{\fancyrefthmlabelprefix}{Theorem~#1}
\frefformat{vario}{\fancyreflemlabelprefix}{Lemma~#1}
\frefformat{vario}{\fancyrefcorlabelprefix}{Corollary~#1}
\frefformat{vario}{\fancyrefdeflabelprefix}{Definition~#1}
\frefformat{vario}{\fancyrefobslabelprefix}{Observation~#1}
\frefformat{vario}{\fancyrefasmlabelprefix}{Assumption~#1}
\frefformat{vario}{\fancyrefasmslabelprefix}{Assumptions~#1}
\frefformat{vario}{\fancyreffiglabelprefix}{Figure~#1}
\frefformat{vario}{\fancyrefapplabelprefix}{Appendix~#1} 
\frefformat{vario}{\fancyrefproplabelprefix}{Proposition~#1}
\frefformat{vario}{\fancyrefalglabelprefix}{Algorithm~#1}
\frefformat{vario}{\fancyrefeqlabelprefix}{(#1)}
\frefformat{vario}{\fancyreftbllabelprefix}{Table~#1}
\frefformat{vario}{\fancyrefestilabelprefix}{Estimator~#1}
\safemath{\dictab}{[\,\dicta\,\,\dictb\,]}

\safemath{\ysig}{\bmy}
\safemath{\ysighat}{\hat{\ysig}}
\safemath{\ysigdim}{M}
\safemath{\xsig}{\bmx}
\safemath{\xsigdim}{N}
\safemath{\nx}{n_x}
\safemath{\zsig}{\bmz}
\safemath{\zsigdim}{\ysigdim}
\safemath{\rsig}{\bmr}
\safemath{\Adict}{\bA}
\safemath{\Adicttilde}{\widetilde{\Adict}}
\safemath{\Adictdim}{\outputdim\times\xsigdim}
\safemath{\avec}{\bma}
\safemath{\avectilde}{\tilde{\avec}}
\safemath{\Bdict}{\bB}
\safemath{\Bdicttilde}{\widetilde{\Bdict}}
\safemath{\Cdict}{\bC}
\safemath{\cvec}{\bmc}
\safemath{\Ddict}{\bD}
\safemath{\Ddictdim}{\ysigdim\times\xsigdim}
\safemath{\dvec}{\bmd}
\safemath{\Ddicttilde}{\widetilde{\bD}}
\safemath{\Bonb}{\bB}
\safemath{\bvec}{\bmb}
\safemath{\Bonbdim}{\ysigdim\times\ysigdim}
\safemath{\noise}{\bmn}
\safemath{\noisedim}{\ysigim}
\safemath{\err}{\bme}
\safemath{\errdim}{\ysigdim}
\safemath{\errset}{\setE}
\safemath{\nerr}{n_e}
\safemath{\delop}{\bP_\errset}
\safemath{\delopc}{\bP_{{\errset}^c}}

\safemath{\cplxi}{\imath}
\safemath{\cplxj}{\jmath}

\safemath{\dict}{\matD}
\safemath{\inputdim}{N}		
\safemath{\outputdim}{M}		
\safemath{\sparsity}{S}	
\safemath{\inputdimA}{{N_a}}	
\safemath{\inputdimB}{{N_b}}	
\safemath{\elemA}{{n_a}}	
\safemath{\elemB}{{n_b}}	
\safemath{\resA}{\matR_a}	
\safemath{\resB}{\matR_b}	
\safemath{\subD}{\matS} 
\safemath{\subA}{\matS_a} 
\safemath{\subB}{\matS_b} 
\safemath{\dicta}{\matA} 	
\safemath{\dictb}{\matB} 	
\safemath{\hollowS}{H}
\safemath{\hollowA}{H_a}
\safemath{\hollowB}{H_b}
\safemath{\cross}{Z}
\safemath{\coh}{\mu_d}			
\safemath{\coha}{\mu_a}			
\safemath{\cohb}{\mu_b}			
\safemath{\mubs}{\nu}	
\safemath{\cohm}{\mu_m} 
\safemath{\dictset}{\setD}	
\safemath{\dictsetp}{\dictset(\coh,\coha,\cohb)}	
\safemath{\dictsetgen}{\dictset_\text{gen}}
\safemath{\dictsetgenp}{\dictsetgen(\coh)}
\safemath{\dictsetonb}{\dictset_\text{onb}}
\safemath{\dictsetonbp}{\dictsetonb(\coh)}

\safemath{\leftside}{U}
\safemath{\rightsideA}{R_a}
\safemath{\rightsideB}{R_b}

\safemath{\indexS}{\setI_S} 

\safemath{\na}{n_a}			
\safemath{\nb}{n_b}			
\safemath{\coeffa}{p_i}	
\safemath{\coeffb}{q_j}	
\safemath{\seta}{\setP}		
\safemath{\setb}{\setQ}     
\safemath{\setw}{\setW}	
\safemath{\setz}{\setZ}	
\safemath{\cola}{\veca}		
\safemath{\colb}{\vecb}		
\safemath{\cold}{\vecd}		
\safemath{\inputvec}{\vecx} 	
\safemath{\error}{\vece}	
\safemath{\noiseout}{\vecz} 	
\safemath{\inputvecel}{x}
\safemath{\inputveca}{\vecx_a}
\safemath{\inputvecb}{\vecx_b}
\safemath{\outputvec}{\vecy}	
\safemath{\lambdamin}{\lambda_{\mathrm{min}}}

\safemath{\elltwo}{\ell_2}
\safemath{\ellone}{\ell_1}
\safemath{\ellzero}{\ell_0}
\safemath{\ellinf}{\ell_\infty}
\safemath{\ellinftilde}{\ell_{\widetilde\infty}}
\safemath{\licard}{Z(\coh,\coha,\cohb)}
\safemath{\xsol}{\hat{x}}
\safemath{\xbord}{x_b}		
\safemath{\xstat}{x_s}		
\safemath{\xstatLone}{\tilde{x}_s}
\safemath{\order}{\mathcal{O}} 
\safemath{\scales}{\Theta} 
\safemath{\ones}{\mathbf{1}} 
\safemath{\zeroes}{\mathbf{0}} 
\safemath{\thlone}{\kappa(\coh,\cohb)} 
\safemath{\constoneA}{\delta} 
\safemath{\constoneB}{\epsilon} 
\safemath{\nlarge}{L}				   
\safemath{\sumlarge}{S_\nlarge}
\safemath{\maxlarger}{P_\nlarge}	   
\safemath{\Pzero}{\textrm{P0}}	
\safemath{\Pone}{\textrm{P1}}
\safemath{\vecfir}{\vecw}			 
\safemath{\vecsec}{\vecz}
\safemath{\elvecfir}{w}              
\safemath{\elvecsec}{z}				 
\safemath{\nlargefir}{n}
\safemath{\normout}{\gamma}
\safemath{\auxfun}{h}
\safemath{\supp}{\textrm{supp}}

\safemath{\indexa}{\ell}
\safemath{\indexb}{r}
\safemath{\indexc}{i}
\safemath{\indexd}{j}

\safemath{\project}{P}

\allowdisplaybreaks

\usepackage{color}
\safemath{\Tran}{\textnormal{T}}

\newcolumntype{P}[1]{>{\centering\arraybackslash}p{#1}}

\title{\LARGE \bf Multi-Agent Active Search using Realistic Depth-Aware Noise Model}
	
\author{Ramina Ghods$^{1}$, William J. Durkin$^{2}$, Jeff Schneider$^{1}$
\thanks{$^{1}$R. Ghods and J. Schneider are with the Robotics Institute, School of Computer Science, Carnegie Mellon University, 
	Pittsburgh, PA 15213
	{\tt\small \{rghods, schneide\}@cs.cmu.edu} }
\thanks{$^{2}$W. Durkin is with the School of Earth Sciences, Ohio State University, Columbus, OH 43210
{\tt\small durkin.98@osu.edu} }}

\begin{document}

\maketitle

\begin{abstract}


The active search for objects of interest in an unknown environment has many robotics applications including search and rescue, detecting gas leaks or locating animal poachers. Existing algorithms often prioritize the location accuracy of objects of interest while other practical issues such as the reliability of object detection as a function of distance and lines of sight remain largely ignored. 
Additionally, in many active search scenarios, communication infrastructure may be unreliable or unestablished, making centralized control of multiple agents impractical. We present an algorithm called Noise-Aware Thompson Sampling (NATS) that addresses these issues for multiple ground-based robots performing active search considering two sources of sensory information from monocular optical imagery and depth maps. By utilizing Thompson Sampling, NATS allows for decentralized coordination among multiple agents. NATS also considers object detection uncertainty from depth as well as environmental occlusions and operates while remaining agnostic of the number of objects of interest. Using simulation results, we show that NATS significantly outperforms existing methods such as information-greedy policies or exhaustive search. We demonstrate the real-world viability of NATS using a pseudo-realistic environment created in the Unreal Engine 4 game development platform with the AirSim plugin.
\end{abstract}




\section{Introduction}
\label{sec:introduction}

Active search (active sensing) refers to the problem of locating targets in an unknown environment by actively making data-collection decisions and finds use in many robotics applications such as search and rescue, localization and target detection \cite{murphy2004human,ma2017active,ghods2020asynchronous,jennings1997cooperative}.
While there is a large amount of research on localization and detection algorithms in robotics, the majority of these algorithms are simplified 
and do not consider the practical side of fielding real sensors such as applying real detection to their observations.  For example, a basic SLAM (simultaneous localization and mapping) focuses on the uncertainty of locations while abstracting the detection of objects \cite{huang2019survey,leonard1991simultaneous}.  Similarly, common coverage planners produce only simplistic plans with an abstraction on the detectors \cite{galceran2013survey}.
Field of search theory does consider uncertainty measures of false positive and false negative in their object detection \cite{kress2008optimal,chung2011analysis}. However, they assume simplified point-wise sensing actions that do not support typical field sensor setups that use common cameras paired with detectors.
Active learning methods such as adaptive compressed sensing \cite{haupt2009adaptive,malloy2014near}, Bayesian optimization \cite{rajan2015bayesian,marchant2012bayesian} and bandit-style algorithms \cite{abbasi2012online,carpentier2012bandit} contain sophisticated reasoning about uncertainty but use simplified sensing models.


Besides sensor uncertainties, executing active search with multiple robots adds an additional challenge.  
While centralized planning is one approach to multi-agent settings, it is often impractical due to communication constraints highly discussed in robotics (\cite{murphy2004human,feddema2002decentralized,yan2013survey,robin2016multi}).
%
Essentially, a central coordinator that expects synchronicity from all robots is not feasible as any communication or agent failure could disrupt the entire process. To clarify, there must be at least some communication between agents to share information, otherwise they are just independent actors, not a team.  We will assume agents do communicate their acquired measurements, yet each agent independently decides on its next sensing action using whatever information it happens to receive.

All these challenges are motivated by real world requirements such as those used by the multi-robot search team in \cite{NREC}
which includes decentralized autonomy in perception, navigation and path planning. However, they require an operator to dictate waypoints (goal locations) to robots. In this paper, we focus on developing an autonomous decentralized multi-agent active search method that performs waypoint selection while taking into account practical field sensors.

To consider practical field sensors, we propose quantitatively modeling their behaviors as follows. 
When a real, autonomous robot performs object detection on a sensed image, 
it reports detections probabilistically and its precision-recall curves degrade with distance between the object and the detector (depth). The robot's performance is also constrained by the field of view of the device as well as occlusions created by terrain or other obstacles in the scene. By modeling all these features, we can expect an efficient algorithm that will start by choosing actions that offer wide views over great distances and will then need to consider the best way to follow up on locations with uncertain detections. 

To develop a decentralized multi-agent algorithm, we propose using parallelized Asynchronous Thompson Sampling \cite{kandasamy2018parallelised}. Thompson Sampling (TS) is an online optimization method 
that balances between exploration and exploitation by maximizing the expected reward 
assuming that a sample from the posterior is the true state of the world \cite{thompson1933likelihood,russo2018tutorial}. TS is an excellent candidate for an asynchronous multi-agent online algorithm without a central planner.
Essentially, by using a posterior sample in its reward function, TS allows a calculated randomness in the reward that enables multiple agents to independently solve for different values that equally contribute to the overall goal of locating targets.

\subsection{Contributions}
\begin{itemize}
	\item We propose a novel multi-agent active search algorithm called NATS (Noise Aware Thompson Sampling) that actively locates sparse targets in an unknown
	environment where agents make independent data-collection decisions asynchronously and in a decentralized manner.

\item 
NATS deploys the practical considerations of real sensors including object detector's uncertainty increase with depth, sensors' field of view and terrain occlusions.
\item We provide simulation results showing that NATS significantly outperforms existing methods such as exhaustive search or information-greedy policies. Using simulations, we also show that NATS is efficient and its complexity is not affected by number of agents or targets.
\item We further demonstrate our algorithm using the Unreal Engine game development platform with the Airsim plugin. We model object detection depth uncertainty, terrain occlusions and field of view in the development of NATS in this game platform. Using this platform, we also provide a valuable dataset which demonstrates an object detector's uncertainty increase with depth.
\end{itemize}



\subsection{Related Work}\label{sec:RW}

Active search closely falls under the category of information gathering in robotics. Originally, much of the work in this field had their focus on single-agent settings (\cite{cliff2015online,lim2016adaptive,patten2018monte,arora2019multi}), or if they were multi-agent, they required a central planner (\cite{cho2018informative,charrow2014cooperative,surmann2019integration}). Recently there has been more attention towards the need for decentralized solutions especially using planning methods \cite{queralta2020collaborative,zhang2019multi}.
%
For example,
\cite{lauri2020multi} uses partially observed Markov decision processes where
agents only have access to their own observations.  Planning is done centrally in advance, and evidence fusion
is done centrally at the end. Only the execution is decentralized.
%
In multi-agent reinforcement learning, \cite{lowe2017multi} and \cite{gupta2017cooperative} make a similar assumption to execute their centrally learned policy in a decentralized manner.
\cite{best2019dec,best2020decentralised} achieve decentralization by having agents repeatedly communicate their future plans with each other, while \cite{hollinger2009efficient} and \cite{dames2017detecting} communicate their future plans in sequence.
In contrast, our problem setting is motivated by a real multi-robot system 
with unreliable communication \cite{NREC}, where we want to benefit from observation 
sharing when it occurs, but never depend on communication for coordination.
While a few of the work mentioned above (e.g. \cite{charrow2014cooperative}) use location uncertainty in their modeling, \cite{dames2017detecting} is the only work we have seen in this field that considers existence uncertainty based on distance in their sequential process.

%
Another solution to active search is the use of entropy and information (\cite{rajan2015bayesian,jedynak2012twenty,ma2017active}). Adapting these solutions to decentralized settings and without the permission to share future plans can be challenging. This is because they are deterministic methods which cause each agent to choose the same action, thus resulting in poor performance unless some stochasticity or coordination mechanism is added. We show this performance problem in our empirical comparison.



%
In search theory, \cite{kress2008optimal} and \cite{chung2011analysis}
consider existence uncertainty through false positive detection but do not provide any insight on how it is related to sensor capabilities.
Additionally, search theory is designed for single-cell sensing and cannot be extended to decentralized multi-agent settings \cite{kriheli2016optimal,asfora2020mixed}.
Active search has a different goal than SLAM since it assumes that robot localization is provided by SLAM and is locating objects of interest. 
%
Nonetheless, bringing our attention to how SLAM literature manages sensor uncertainty, we see that
they generally consider the uncertainty of the \textit{location} of features but are not concerned with uncertainty in their existence \cite{davison2007monoslam,engel2014lsd,mur2015orb}.
%
%
%
%
The computer vision literature also contains significant work on
uncertainty of detection and location of object pixels within an image
(e.g. \cite{hall2020probabilistic,kampffmeyer2016semantic,gonzalez2015active,caicedo2015active}). However, we are interested in the problem of efficiently choosing images to physically locate these objects.

\cite{reid2013cooperative,olson2012progress} are part of a contest in 2010 that closely matches our problem settings.  
However, they performed object detection as an extension of
SLAM rather than developing efficient algorithms to find objects in
the face of uncertain sensing.
In semantic mapping, \cite{liu2014extracting} represents uncertainty in the existence of features but their driving goal is localization, not efficient discovery of objects of interest.
Other areas of work include search and pursuit-evasion \cite{chung2011search} or next best view \cite{lauri2020nbv} which are not the focus of this paper. 
\cite{qingqingtowards} is a recent work that provides a dataset on how altitude affects object detection. However, they are not mathematically modeling their findings nor do they use it for any algorithm.

In \cite{ghods2020asynchronous}, we proposed a parallelized Thompson Sampling algorithm for active search called SPATS that allows agents to make independent and intelligent decisions in a decentralized manner. However, SPATS is limited to a simplified sensing action that entirely ignores the presence and effects of object detection and is more suited for unmanned aerial vehicles.

\vspace*{-0.4mm}

\subsection{Notation}
Lowercase and uppercase boldface letters represent column vectors and matrices, respectively. For a matrix $\bA$, the transpose is  $\bA^T$.
For a vector~$\bma$, the $i$th entry is~$[\bma]_i$ or $a_i$.
%
The $\ell_2$-norm of~$\bma$ is~$\|\bma\|_2$; $\diag(\bma)$ is a square matrix with $\bma$ on the main diagonal,
and the trace operator is $\tr(\cdot)$. 
%
%
%


\section{Problem Formulation and Sensing Model}
\label{sec:formulation}

\subsection{Problem Definition}

\begin{figure}
	\centering
	\vspace*{2mm}
	\begin{subfigure}{0.49\linewidth}
		\centering
		\includegraphics[width=0.85\linewidth]{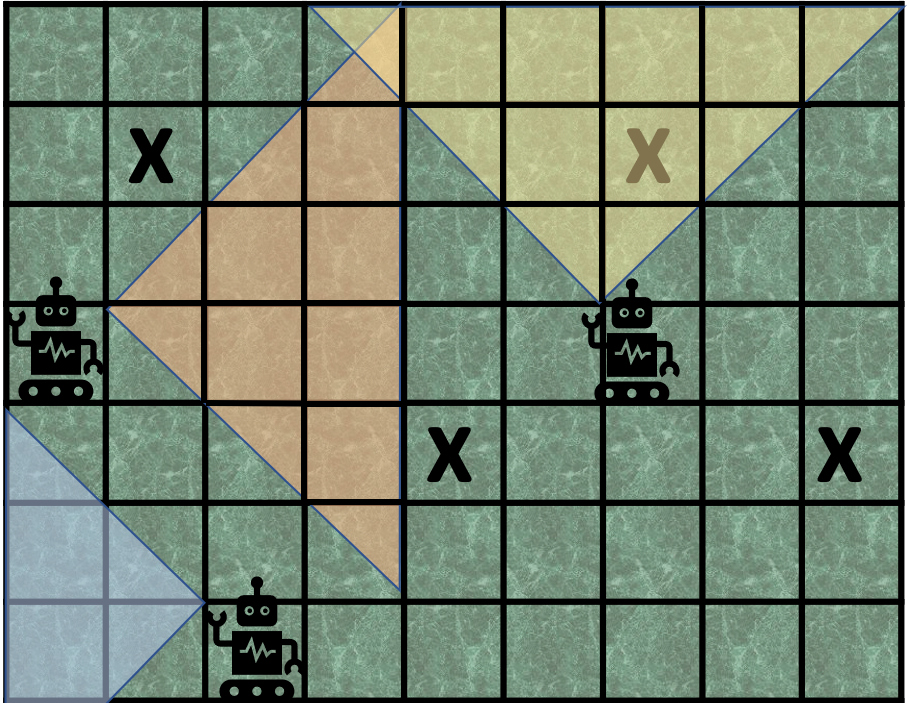}
		\caption{Multi-Agent Active search}
		\label{fig:formulation}
	\end{subfigure}
	\begin{subfigure}{0.49\linewidth}
		\centering
		\includegraphics[width=\linewidth]{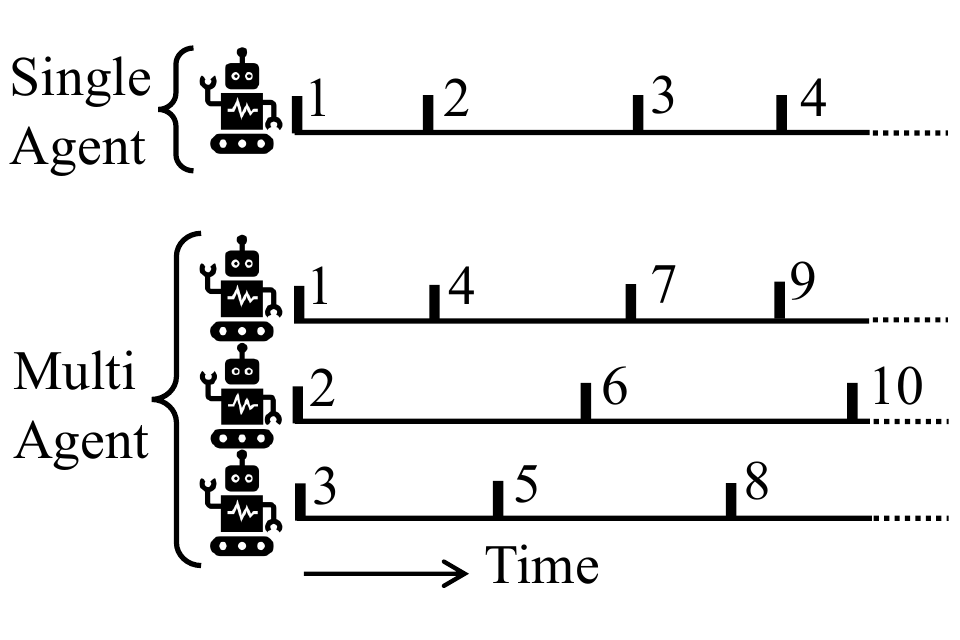}
		\caption{Single vs. Multi}
		\label{fig:asynchronous}
	\end{subfigure}
	\caption{(a) Robots are locating objects of interest by searching the environment using optical images analysed with object detectors. (b) Single-agent vs asynchronous multi-agent. Here, the small numbered horizontal lines indicate the start of $t$'th task. In single agent, tasks start sequentially. In multi
		agent, task $t$ can start before all previous $t-1$ tasks are finished.}
\end{figure}

Consider the gridded area in \fref{fig:formulation} to be an area of interest for active search where the marks “X” show the location of the objects of interest (OOI). Multiple ground robots are sent to the environment to locate said OOIs as fast as possible. Each robot moves around and senses its surrounding by taking pictures and passing them through an object detector, e.g. YOLOv3 \cite{redmon2018yolov3}. The colored triangles in this figure illustrate each robot's sensing action, i.e. $90^o$ field of view (FOV) of their captured image.
Since there is no central planner, each robot must independently decide on their next sensing action given their current belief on the location of the OOIs.
%
%

Once a robot senses a region, it will run the sensed image through an object detector that will extract OOIs with a given confidence score. In general, objects farther away from the camera will have a lower probability of being correctly identified. We can measure this probability for a given object detector using training data. Our objective is to model this probability as a function of the object's distance from the camera and utilize it to help the robot with making sensing decisions. In particular whether or not the robot should give a closer look to a region that is likely to include an OOI. We will provide a model for this probability in the next section.

\begin{figure}[t]
			\vspace*{2mm}
	\centering
	\includegraphics[width=\linewidth]{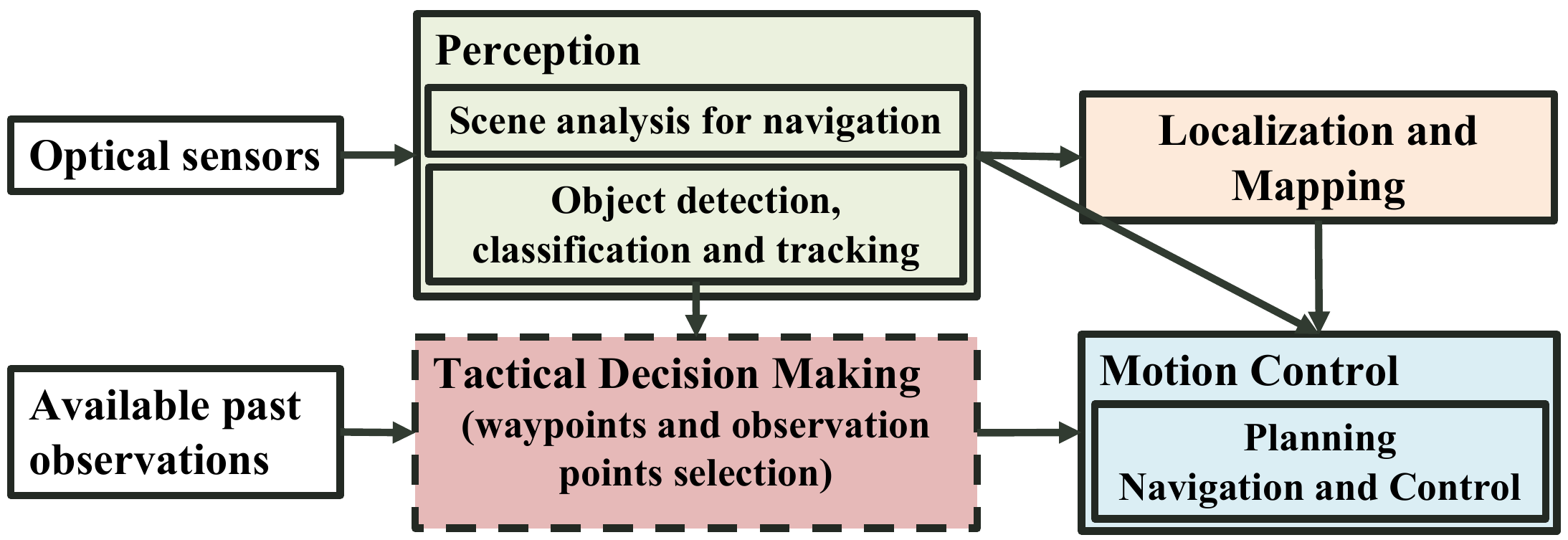}
	\caption{Basic simplified architecture of an autonomous robot. Our algorithm is focusing on Tactical Decision Making.}
	\label{fig:architecture}
\end{figure}

We note that we are not replacing the localization and mapping portion of robot autonomy. We assume the robots are able to localize themselves. Our goal is to make the tactical decisions on the next waypoints (sensing action) at each time step. In particular, we use \fref{fig:architecture} to illustrate a simplified architecture of an autonomous robot. Our objective is to develop an algorithm for the dashed red box. To simplify the problem setting, we assume the robots only sense at goal locations and not while they are travelling between them.
%

\vspace*{1mm}
\hspace*{-3.5mm}\textbf{Communication Setup:}
Despite unreliability, we do assume communication will be available sometimes and want to take 
advantage of it when possible.  That leads to the
following constraints for our algorithm:
1) The agents share their past actions and observations when possible.
2) There can be no requirement that the set of available past measurements remains consistent across agents since communication problems
can prevent it.
3) There can be no part of the algorithm where
an agent must wait for communication from its teammates
before acting since this wait could be arbitrarily long and
thus cause a loss of valuable sensing time.

\subsection{Depth-Aware Modeling of an Object Detector} \label{sec:ImperfectObjDet}
We intend to formulate the performance of an object detector with an additive noise model. Let us assume $\beta_i$ is the output of an ideal object detector that identifies object $i$ that is a distance $\ell_i$ away from the camera with either a ``0'' (not OOI) or a ``1'' (OOI). An imperfect object detector can sometimes misclassify the OOI with a false positive or a false negative. Therefore, one way to model the performance of the object detector is to model the misclassifications with an appropriate noise distribution such as Gaussian distribution with its variance describing the false positive and false negative rate. 
%
While this model is reasonable, it is disregarding an important piece of information on the confidence score provided by the object detector. 
In general, when the object detector makes a mistake, we expect it to generate a lower confidence score \cite{jiao2019survey}. In fact, we make the following claim:

\vspace*{1mm}
\hspace*{-3.5mm}\textbf{Claim 1.}\hspace{1mm}
\textit{We expect the confidence score of an object detector to gradually decline as a function of the object's distance from the camera (assuming fixed focal length).}

In \fref{sec:results}, we provide a dataset to back up this claim for YOLOv3 using images from a realistic environment we have created in Unreal Engine. 
Note that Claim 1 is not considering active vision through camera zooming or gimbal stabilization as they are not the focus of this work \cite{konda2014real,warren2018level}.

\begin{figure}[t]
	\vspace*{2mm}
	\centering
	\includegraphics[width=0.6\linewidth]{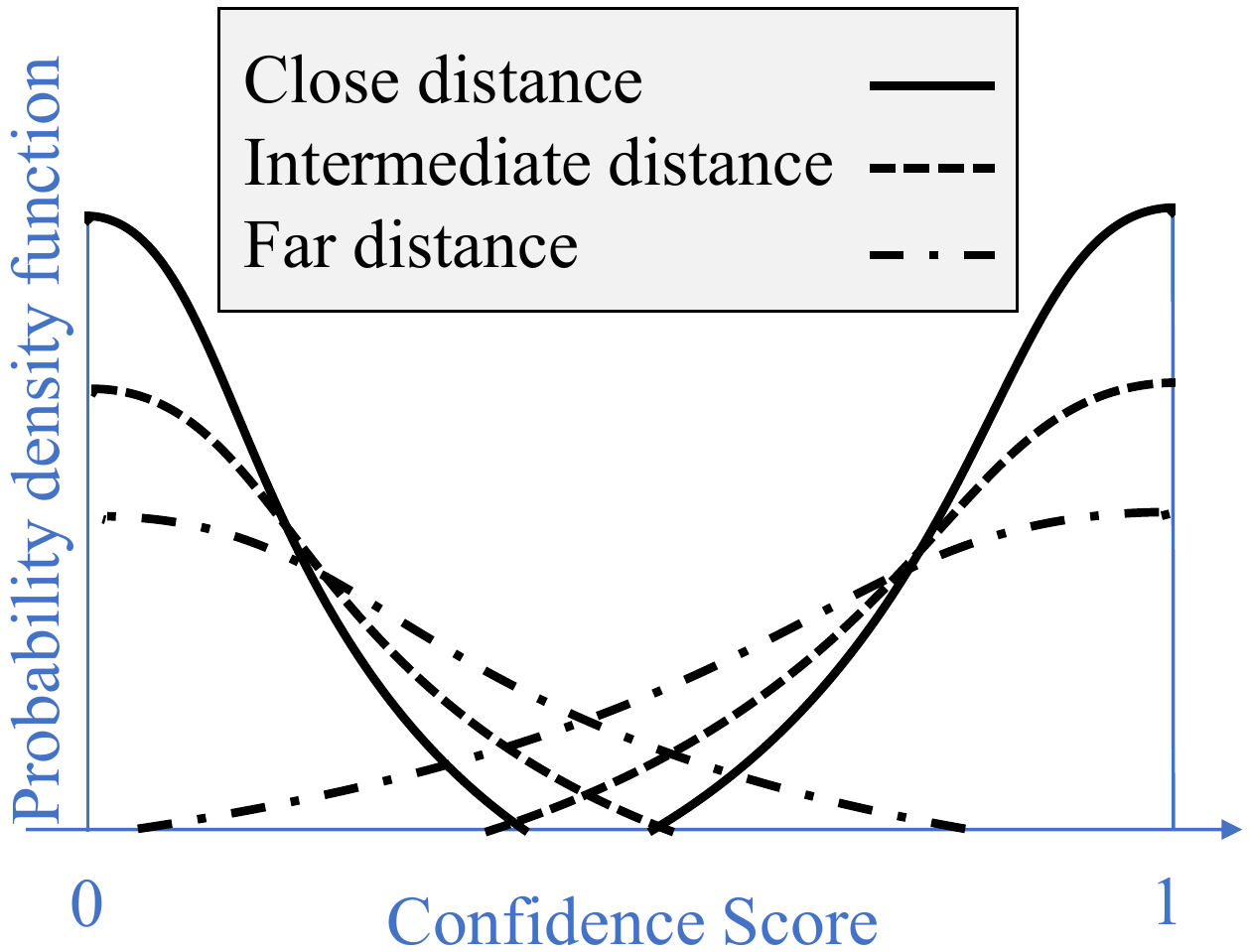}
	\caption{Probability density function of an object detector's confidence score for different object distances.}
	\label{fig:AccDistModel}
\end{figure}

Using Claim 1, we model the performance of an imperfect object detector by formulating its confidence score ($y_i$) with an additive one-sided Gaussian noise as depicted in \fref{fig:AccDistModel}. Precisely, for any given distance $\ell_i$, we have
$y_i \!=\! \beta_i \!+\!n_i,$ with $n_i \!\sim\! \setN^+\!(0,\sigma_{i}^2(\ell_i)\!)$.
Here, the variance $\sigma_{i}^2(\ell_i)$ is an increasing function of $\ell_i$ that can be computed using training data.

\subsection{Problem Formulation}
We describe the mission environment (the gridded environment in \fref{fig:formulation}) with a sparse matrix $\bB \in \reals^{M_1 \!\times \! M_2}$ with $k$ non-zero entries at the location of OOIs. We consider each entry of $\bB$ to be the output of a perfect object detector with confidence score 0 for ``no OOI'' and score 1 for ``OOI''. Defining $\bm\beta \in \reals^{M \!\times\! 1}$ as a flattened version of matrix $\bB$ with $M\!=\!M_1 M_2$,
we write the sensing operation for each agent at time $t$ as:
\vspace*{-3mm}
\begin{align}\label{eq:formulation}
\,\,\,\,\,\,\,\,\,\,\,\,\,\,\,\, \bmy_t=\bX_t \bm\beta+\bmn_t, \quad \bmn_t \sim \setN^+(0,\bSigma_t).
\end{align}
Here, matrix $\bX_t \in \reals^{Q \times M}$ describes the sensing matrix at time t (colored triangles representing the robot's FOV). To better describe the sensing matrix $\bX_t$, consider \fref{fig:formulation}. Essentially, each row of the sensing matrix $\bX_t$ is a one-hot vector pointing out the location of one of the colored grid points inside the robot FOV triangle. We will discard entries of this FOV grids that are unavailable due to occlusion. We assume there are $Q$ FOV grid points available at time step $t$. 

Next, $\bmy_t \in \reals^{Q \times 1}$ is the observation vector modeling the output of an imperfect object detector. 
$\bmn_t \in \reals^{Q \times 1}$ is a depth-aware additive noise vector where each of its entries $[\bmn_t]_q \sim \setN^+(0,\sigma_{q}^2(\ell_q))$ ($q\!=\!1,..,Q$)  is modeling the noise from the imperfect object detector defined in \fref{sec:ImperfectObjDet}. Specifically, For each of the $Q$ grid points in the robot's FOV, we consider its observations $[\bmy_t]_1,...,[\bmy_t]_Q$ to be corrupted with independent additive Gaussian noises $[\bmn_t]_1,...,[\bmn_t]_Q$. The variance $\sigma_{q}^2(\ell_q)$ of each noise entry is a function of the distance between grid point index $q$ and the robot ($\ell_q$). We define the noise variance $\bSigma_t$ as a diagonal matrix with each of its entries referring to the noise variance for the corresponding FOV grid points, i.e. $\bSigma_t=\text{diag}(\sigma_{1}^2(\ell_1),...,\sigma_{Q}^2(\ell_Q))$.

\vspace*{1mm}
\hspace*{-3.5mm}\textbf{Remark 1.}
	\textit{
	Note that since the focus of our algorithm is to provide goal locations for each agent (not to plan a continuous path), we only need a very coarse discretization for our environment. For example, in \fref{sec:results} we will use grid sizes of $30 \!\times\! 30$m to cover a $500 \!\times\! 500$m area.
}

\vspace*{1mm}
\hspace*{-3.5mm}\textbf{Remark 2.}
\textit{Note that while we only allow one OOI in each grid point, it is easy to modify the sensing model in \fref{sec:ImperfectObjDet} (allow values between ``$0$'' and ``$2$'' for $y_i$) to estimate multiple OOIs in the same grid.}

To best estimate $\bm\beta$ and actively locate OOIs, at each time step $t$, agent $j$ choose a sensing action $\bX_t$ given all the available measurements thus far in its measurement set $\bD^j_{t}$. Let us assume the collective number of measurements available to all agents are $T$. Our main objective is to correctly estimate the sparse vector $\bm\beta$ with as few measurements $T$ as possible. 
We also expect the agents to achieve this objective with a short travelling distance (see last of \fref{sec:NATS}).

For a single agent the action selection process is sequential with the measurement sequence $\bD^1_{t}=\{(\bX_1,\bmy_1 ),...,(\bX_{t-1},\bmy_{t-1})\}$ available to the agent at time step $t$. For a multi-agent setting, we use an asynchronous parallel approach with multiple agents independently making data-collection decisions
as proposed in \cite{kandasamy2018parallelised,ghods2020asynchronous}. Precisely, as illustrated in \fref{fig:asynchronous} the asynchronicity means that the agents will not wait on results from other agents;
instead, an agent starts a new sensing action immediately after its previous data acquisition is completed using all the measurements available thus far. As an example, the second agent ($j=2$) in the multi-agent example in \fref{fig:asynchronous} will start task 6 before tasks 4 and 5 are completed with $\bD^2_{6}=\{(\bX_{t'},\bmy_{t'}) | t'=\{1,2,3\}\}$.

%




\section{Our Proposed Algorithm: NATS}\label{sec:LTS}

\subsection{Roadmap to NATS}\label{sec:sparseTS}

As detailed in the introduction, Thompson Sampling (TS) is a great solution for our decentralized problem. In particular, TS introduces a randomness from posterior sampling in each agent's reward function that allows them to independently contribute to the overall goal of locating OOIs. We will next discuss two steps of adapting TS to our active search problem.

First, we realize that our active search falls in the category of parameter estimation in active learning as developed in \cite{kandasamy2019myopic} with the name Myopic Posterior Sampling (MPS). Similar to MPS, our goal is to actively learn parameter $\bm\beta$ by taking as few measurements as possible. 
For the sake of similarity, we use TS to refer to MPS. 
Here, we choose a myopic solution (one-step lookahead) over a planning approach due to computational overhead. While applying planning methods in low dimensional or fully observable environments is very effective (e.g. \cite{sartoretti2019distributed,flaspohler2019information}), our model has high dimensions, has sparse rewards and includes multiple levels of uncertainty in its formulation. Keeping track of all these uncertainties and high sample dimensions in a planning approach (e.g. Monte Carlo Tree Search) can become computationally intractable. 

Our second step to adapt TS is choosing an inference method. To perform active search, traditionally people have used coverage planning methods with exhaustive search (\cite{lin2009uav,chien2010human,ryan2005mode}). 
However, with the availability of observations with high and low uncertainty, an optimized active search method can locate OOIs faster than exhaustive search in terms of number of measurements (see \fref{sec:results} for examples). Such faster recovery is achievable due to the concept of sparse signal recovery which says that we can recover a sparse signal with size $M$ by taking less than $M$ linear measurements \cite{candes2006robust,donoho2006compressed}. By using sparse signal recovery as the inference method for TS, we can create the right balance between exploring unknown regions with large uncertainty and then exploiting the ones we suspect of including an OOI with a closer look.

\subsection{Developing NATS (Noise-Aware Thompson Sampling)}\label{sec:NATS}
We will now derive the TS algorithm with sparse prior for each agent. Once agent $j$ finishes an observation task, it will use all the available measurements to it at that point ($\bD^j_{t}$) to start a new sensing operation. We can divide this process into two stages of posterior sampling and design as follows.

%
\subsubsection{Posterior Sampling Stage}
Recall our interest in estimating parameter $\bm{\beta}$ in \fref{eq:formulation}. In this stage, agent $j$ computes a posterior belief for $\bm\beta$ and takes a sample from it.
 Assuming a prior $p_0(\bm{\beta})$ and given the likelihood function $p(\bmy_{t'} | \bX_{t'},\bm\beta) = \setN(\bX_{t'} \bm\beta ,\bSigma_{t'})$ for all previous measurements $(\bX_{t'},\bmy_{t'}) \in \bD^j_{t}$, we can compute the posterior distribution as
\begin{align}\label{eq:posterior}
\textstyle	p(\bm\beta | \bD^j_{t}) = \frac{1}{Z} p_0(\bm\beta)\prod_{(\bX_{t'},\bmy_{t'}) \in \bD^j_{t}} p(\bmy_{t'} | \bX_{t'}, \bm\beta).
\end{align}

Now, we need to choose a prior distribution $p_0(.)$ for the vector $\bm\beta$ to compute its corresponding posterior. Since $\bm\beta$ is sparse with an unknown number of non-zeros, we use sparse Bayesian learning (SBL) originally proposed by \cite{tipping2001sparse}. We choose SBL for multiple reasons as pointed out by \cite{o2019sparse}. 1) In many cases SBL have shown to achieve more accurate recovery results than $\ell_1$-norm based regularization methods \cite{wipf2009solving,wipf2011latent}. 2) SBL uses a simple Gaussian-based probabilistic model that makes computing the TS reward simpler. 3) SBL allows for automatically tuning the unknown sparsity rate parameter $k$ through an Expectation-Maximization process. We now briefly discuss SBL for our problem setting.

We place a zero-mean Gaussian prior per entry of vector $\bm\beta$ as in $p_0(\beta_m) = \setN(0,\gamma_m)$, with variances $\gamma_m$ as hyperparameters ($m=1,...,M$). Since a Gaussian distribution does not impose sparsity, SBL framework introduces sparsity by choosing variances $\gamma_m$ appropriately given measurements. Essentially, SBL chooses very small values for $\gamma_m$ imposing sparsity unless compelling evidence proves a non-zero entry. 
Using this Gaussian prior along with our Gaussian likelihood $\prod_{(\bX_{t'},\bmy_{t'}) \in \bD^j_{t}} p(\bmy_{t'} | \bX_{t'},\bm\beta)$, the posterior distribution in \fref{eq:posterior} is simply a Gaussian distribution $p(\bm\beta | \bD_{t}^j) = \setN(\bm\mu,\bV)$ with:
\begin{align}\label{eq:postMV}
\bV = \left(\mathbf{\Gamma}^{-1}+\bX^\Tran \bSigma \bX \right)^{-1} \quad \& \quad \bm\mu = \bV \bX^\Tran \bSigma \bmy,
\end{align}
where, $\mathbf{\Gamma} = \diag([\gamma_1,..., \gamma_M])$. Matrices $\bX$ and $\bmy$ are created by vertically stacking all measurements in $(\bX_{t'},\bmy_{t'}) \in \bD^j_{t}$. For example, if $\bD^j_{t}=\{(\bX_{1},\bmy_{1}),(\bX_{2},\bmy_{2})\}$, then $\bmy\!=\![\bmy^\Tran_1,\bmy^\Tran_2]^\Tran, \bX\!=\![\bX_1^\Tran\!,\bX_2^\Tran]^\Tran\!$. Variance $\bSigma$ is a diagonal matrix containing their corresponding depth-aware noise variance. 

Using a conjugate inverse gamma prior for hyperparameters $\gamma_m$ as $p(\gamma_m) = \setI \setG (a_m,b_m) = \frac{b_m^{a_m}}{\Gamma(a_m)}\gamma_m^{(-a_m-1)} e^{-(b_m/\gamma_m)}$, SBL optimizes these parameters by applying an expectation-maximization \cite{tipping2001sparse,wipf2004sparse}. With $\bm\beta$ as the hidden variable, the expectation step follows that of \fref{eq:postMV}, while the maximization step is given by maximizing the likelihood $p(\bmy|\mathbf{\Gamma},\bX)=\int p(\bmy|,\bX,\bm\beta) p(\bm\beta|\mathbf{\Gamma}) \text{d} \bm\beta$ which compiles to:
\begin{align}\label{eq:empar}
\quad \quad \textstyle \gamma_m = {\left([\bV]_{mm}+[\bm\mu]_m^2+2b_m\right)}/{\left(1+2a_m\right)}.
\end{align}

Lastly, agent $j$ samples from the posterior $\tilde{\bm\beta} \sim p(\bm\beta | \bD^j_{t})$ in \fref{eq:postMV} which is very easy due to its Gaussian distribution.

\subsubsection{Design Stage} 
In this stage, agent $j$ chooses sensing action $\bX_{t}$ by maximizing a reward function that assumes the posterior sample $\tilde{\bm\beta}$ is the true beta. Specifically,
assume $\hat{\bm \beta}(\bD^j_{t} \cup (\bX_{t},\bmy_{t}) )$ is our expected estimate of parameter $\bm\beta$ using all available measurements $\bD^j_{t}$ and one-step future measurements $(\bX_{t},\bmy_{t})$. Then, TS will choose future measurements that allow $\hat{\bm \beta}(.)$ to be as close as possible to the posterior sample $\tilde{\bm\beta}$.
In particular, we will use the negative mean square error as our reward function, i.e. $\textstyle \setR(\tilde{\bm\beta},\bD^j_{t},\bX_{t}) = - \Ex{\bmy_{t}|\bX_{t},\tilde{\bm\beta}}{\|\tilde{\bm\beta}-{\hat{\bm \beta}(\bD^j_{t} \cup (\bX_{t},\bmy_{t}) )}\|_2^2}$. 

Using $\bm\mu$ in \fref{eq:postMV} as the posterior mean estimate for ${\hat{\bm \beta}(\bD^j_{t} \cup (\bX_{t},\bmy_{t}) )}$, it is straightforward to compute the reward $\setR(\tilde{\bm\beta},\bD_{t},\bX_{t})$ for TS  at time step $t$ as follows. 
\begin{align}\nonumber
&\textstyle \setR(\tilde{\bm\beta},\!\bD^j_{t},\!\bX_t) \!=\! -\Ex{\bmy_{t}|\bX_{t},\tilde{\bm\beta}}{\|\tilde{\bm\beta}\! - \!\hat{\bm\beta}(\bD_{t}\cup (\bX_{t},\bmy_{t}))\|_2^2}\\\nonumber
&\text{\resizebox{0.85\hsize}{!}{$=\!\textstyle - \| \bV \bX^\Tran \Sigma \bmy \!- \!\tilde{\bm\beta} \|_2^2 \!-\! \|\bV \bX_{t}^\Tran \Sigma_{t}\|_2^2 \! \left(\!\tr(\Sigma_{t}) \!+\! \|\bX_{t} \tilde{\bm\beta}\|_2^2 \!\right) 
\!\!$}}\\\label{eq:reward}
&\text{\resizebox{0.5\hsize}{!}{$\textstyle -2\left(\!\bV \bX^\Tran \Sigma \bmy \!-\! \tilde{\bm\beta}\!\right)^\Tran \!\bV \bX_{t}^\Tran \Sigma_{t} \bX_{t} \tilde{\bm\beta}$}}.
\end{align}
%
%
To maximize the reward above, agent $j$ must choose a feasible action $\bX_{t}$ that represents the FOV of an image captured by a robot. As an example, the colored triangles in \fref{fig:formulation} are feasible actions of robots with $90^o$ FOV.
Given this practical constraint, there is no closed form solution to optimize for this reward.
Consequently, we will have each agent consider a group of feasible sensing actions in a fixed radius surrounding the agent's current location. 
This strategy has two great benefits for us. First, our algorithm is taking into account travelling distance costs for each agent. Second, the size of the environment will not affect the optimization search size. If we do not wish to limit the action set, we could add a simple term of $- \alpha \|\bX_{t-1} \!-\! \bX_{t}\|^2_2$ to the reward in \fref{eq:reward} to account for travelling cost.
 \fref{alg:NATS} summarizes our proposed NATS.
\section{Experimental Results}
\label{sec:results}

\subsection{Synthetic data}
 We now compare NATS against 5 other methods in a synthetic setup. 
1) An information-theoretic approach called ``\emph{RSI}'' proposed in \cite{ma2017active} that we have extended to multi-agent systems. RSI is a single agent active search algorithm that locates sparse targets while taking into account realistic sensing constraints. 2) A TS algorithm similar to NATS that uses a simple Bernoulli prior $p_0(\beta_m=1)=k/M$ and $p_0(\beta_m=0)=1-k/M$. We call this algorithm ``\emph{BinTS}'' (for Binary TS) and assume it has perfect knowledge of sparsity rate $k$. This comparison helps us understand the role of sparsity in our algorithm. 3) ``\emph{Rnd}'' which randomly chooses sensing actions at all times. 4) A point-sensing method we call ``\emph{Point}'' that exhaustively searches the environment. 5) An info-greedy approach we call ``\emph{IG}'' that computes information gain using the negative entropy of the posterior in \fref{eq:postMV}.

Consider an environment with $m_1 \!\times \!m_2 = 16 \!\times\! 16$ grid points. We send $J$ agents over to actively search the environment to recover vector $\bm\beta$ which is randomly generated using a uniform sparse prior with $k$ non-zero entries with value 1. We assume agents can only be placed in the center of these grid points and are free to move between them in any direction. 
For this experiment, we consider the following action set. Each agent can place itself in any of the $16 \!\times\! 16$ grid points in the map. For any given placement, the agent can look in one of $4$ possible directions: north, south, east, or west with $90^o$ FOV. This means that each agent will pick from $16 \times \!16\! \times\! 4$ feasible actions at each time step. In each look-direction, a total of $12$ grid points are visible to the agent: $2$ closest to the agent and $6$ furthest away forming a pyramid shape. 
To consider object detection uncertainty, we assume the sensing actions are affected by three different levels of noise variance ($\{1,4, 9\}\times 0.005$) given their projection distance to the plain parallel to the agent’s location. Algorithms of NATS, BinTS, RSI and IG are all taking these three variances into account.

Here we also assume agents share their measurements with each other soon after they are available. Note that we do not simulate the unreliability of communication here. Rather, we use it to limit communication to past actions and observations. For this synthetic experiment, we consider no travelling cost ($\alpha=0$) and assume there are no occlusions. 



\begin{algorithm}[b]
	\caption{NATS}
	\begin{algorithmic}
		\STATE {\bfseries Assume:} Sensing model \fref{eq:formulation}, sparse signal $\bm\beta$, $J$ agents
		\STATE {\bfseries Set:} $\bD^j_{0} = \varnothing$ ($j=1,...,J$), $\gamma_m = 1$ ($m=1,...,M$)
		\STATE{\bfseries For }{$t=1,...,T$}
		\STATE \hspace{0.4cm}Wait for an agent to finish; for the free agent $j$:
		\STATE \hspace{0.8cm}Sample $\tilde{\bm\beta} \sim p(\bm\beta | \bD^j_{t},\mathbf{\Gamma})= \setN(\bm\mu,\bV)$ from \fref{eq:postMV}
		\STATE \hspace{0.8cm}Select $\bX_{t} \!=\! \argmax_{\tilde{\bX}} \setR(\bm\beta^\star,\bD^j_{t},\tilde{\bX})$ using \fref{eq:reward}
		\STATE \hspace{0.8cm}Observe $\bmy_{t}$ given action $\bX_{t}$
		\STATE \hspace{0.8cm}Update and share $\bD^j_{t+1} \!=\! \bD^j_{t} \!\cup (\bX_{t},\bmy_{t})$
		\STATE \hspace{0.8cm}Estimate $\mathbf{\Gamma}=\diag([\gamma_1,..., \gamma_M])$ using \fref{eq:empar}
	\end{algorithmic} 
	\label{alg:NATS}
\end{algorithm}

\begin{figure}
	\vspace*{2mm}
	\centering
	\begin{subfigure}{0.485\linewidth}
		\includegraphics[width=0.98\linewidth]{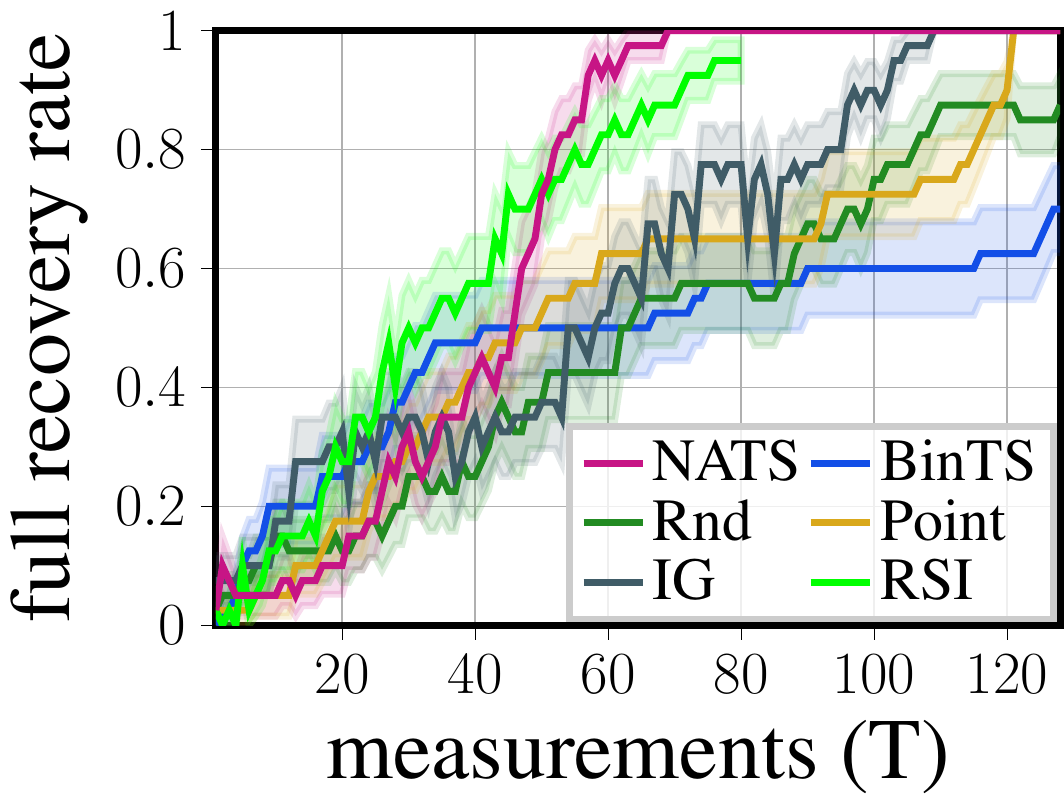}
		\caption{k=1, J=4}
		\label{fig:recovery_k1}
	\end{subfigure}
	\begin{subfigure}{0.485\linewidth}
		\includegraphics[width=0.98\linewidth]{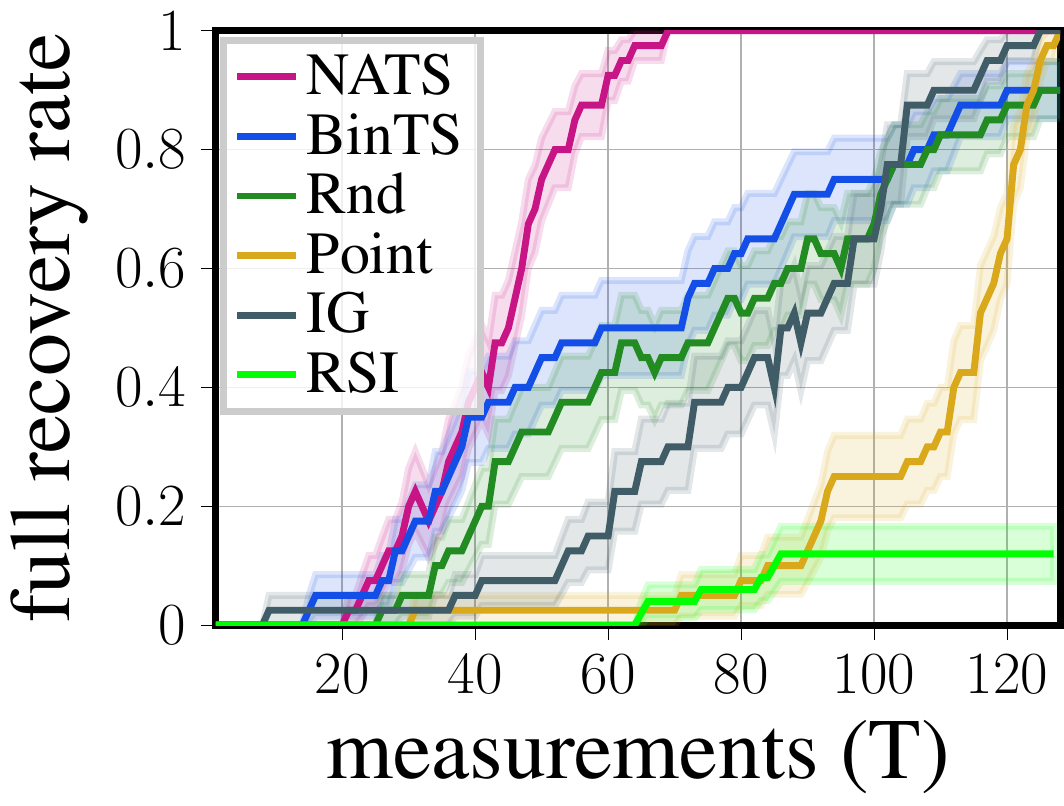}
		\caption{k=5, J=4}
		\label{fig:recovery_k5}
	\end{subfigure}
	\\
	\begin{subfigure}{0.49\linewidth}
		\centering
		\includegraphics[width=\linewidth]{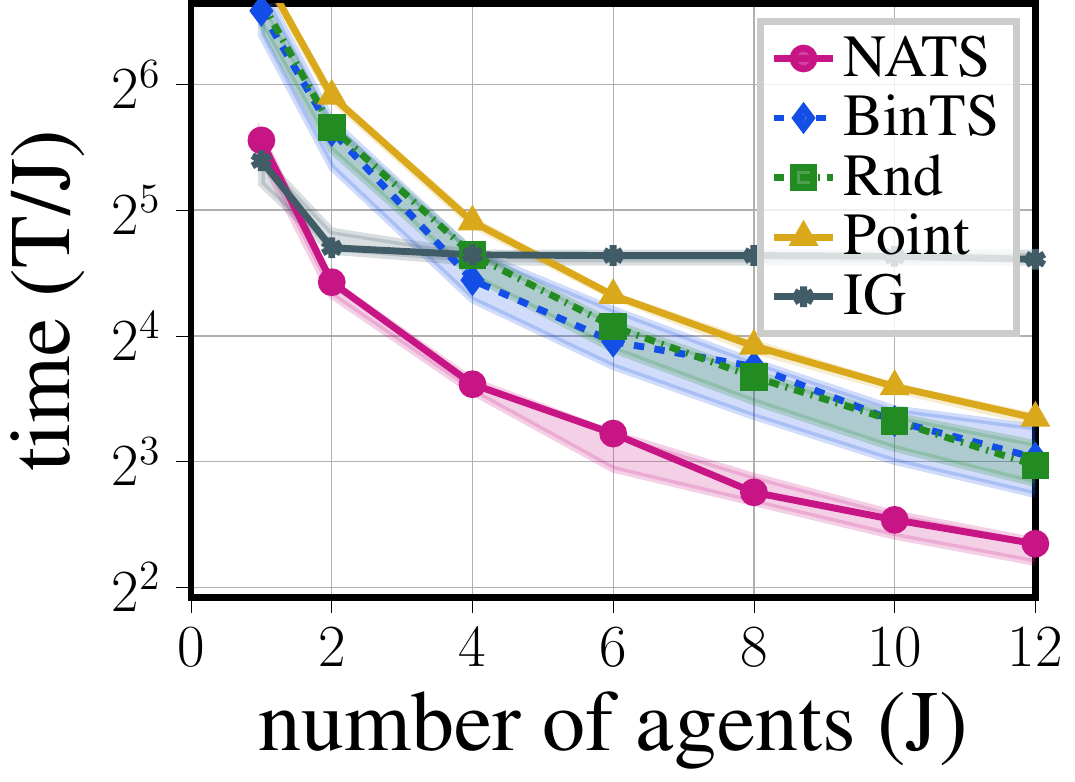}
		\caption{k=5, recovery $\!\ge\! 0.7$}
		\label{fig:agents}
	\end{subfigure}
	\begin{subfigure}{0.49\linewidth}
		\centering
		\includegraphics[width=\linewidth]{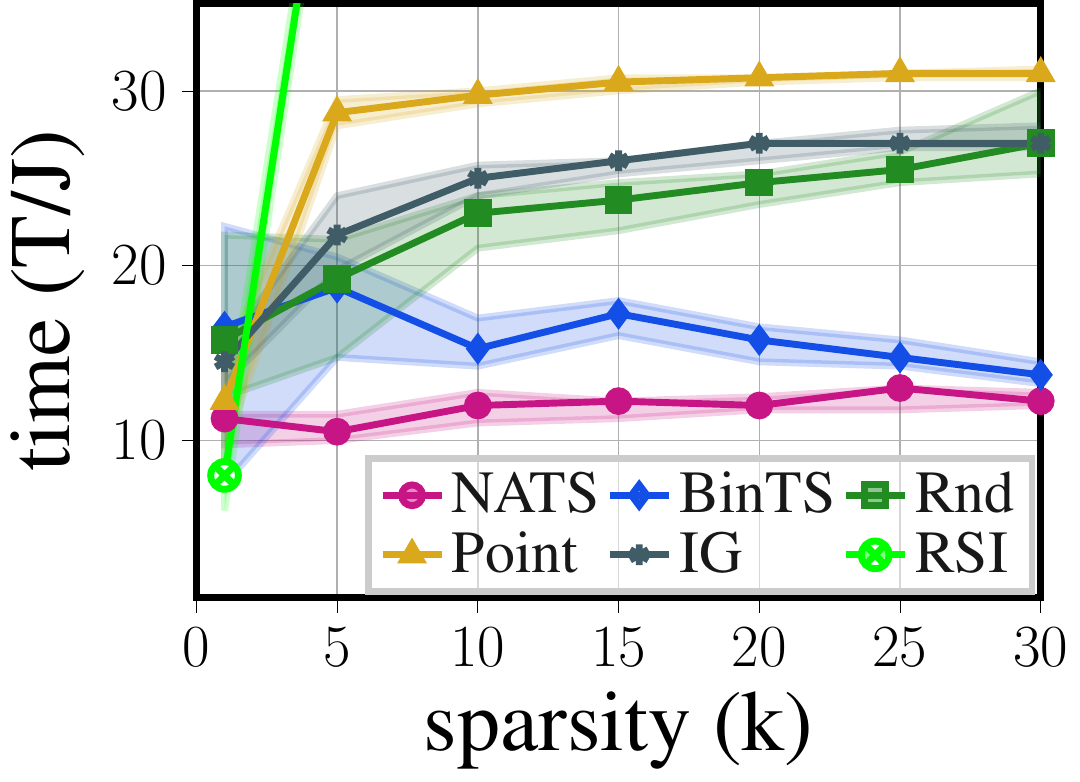}
		\caption{J=4, recovery $\!\ge\! 0.5$}
		\label{fig:sparsity}
	\end{subfigure}
	\caption{Recovery results on synthetic data}
\end{figure}

\fref{fig:recovery_k1} and \fref{fig:recovery_k5} show the results of full recovery rate as a function of number of measurements over random trials. 
In particular, we vary the number of measurements $T$ and compute the mean and standard error of the full recovery rate over $40$ random trials. The full recovery rate is defined as the rate at which an algorithm correctly recovers \emph{the entire vector} $\bm \beta$ over the random trials. Here, $T$ includes the total number of measurements collected by all agents. From these two figures we see that NATS significantly outperforms Point, BinTS, Rnd and IG for both sparsity rates. Here, outperforming BinTS is an evidence on the importance of sparsity that NATS takes into account. Meanwhile, the low performance of IG matches our discussion in \fref{sec:RW} that information-greedy methods result in agents duplicating each other's sensing actions in decentralized multi-agent settings since no randomness or additional coordination is present. In \fref{fig:recovery_k1} with $k=1$, we see that even though RSI is an info-greedy method, its performance is comparable to NATS. 
The reason for this contradicting behavior is that RSI is designed for $k=1$, therefore its performance is so close to optimal (binary search) that it reaches recovery rate of $1$ before the decentralizing can negatively affect it.
For higher sparsity rate of $k=5$, RSI's performance is largely declined. This is a result of poor approximation of mutual information for $k>1$ by RSI and lack of randomness in its reward.
%
Additionally,
RSI uses a sensing model that is not suitable for incorporating object detection confidence scores
and its posterior calculations are highly complex in our simulations.

To further demonstrate the performance of NATS, we provide \fref{fig:agents} and \fref{fig:sparsity}.  \fref{fig:agents}  shows how all methods perform in terms of time as we increase the number of agents for $k=5$. Specifically, we are plotting the time required for each algorithm to reach a minimum full recovery rate of 0.7 for different number of agents $J$. Here, time is defined as the average number of measurements each agent will be collecting in a multi-agent settings, i.e. $T/J$. In an optimal setting, we expect a single agent algorithm's performance to multiply by $J$ as we increase the number of agents. We see that for all algorithms except for IG and RSI, the performance multiplies by $J$ for smaller $J$ values. For NATS, this experiment shows that 
the chances of agents choosing similar actions is very small. As we increase the number of agents beyond 8, the performance improvement reaches incremental levels showing that chances of agents making similar actions are higher as we get closer to maximum performance.  
IG does not improve with agents as without randomness in its reward, all agents are taking the same action. Lastly, since RSI's recovery rate never reached 0.7, its performance plot is excluded from this figure.

%

In \fref{fig:sparsity}, we plot time performance of all algorithms to reach a minimum full recovery rate of 0.5 in terms of sparsity rate $k$. We see here that NATS is a very robust algorithm hardly affected by $k$ (number of OOIs). Rnd has a harder time recovering all OOIs as we increase $k$. As BinTS is designed for non-sparse vectors, its performance improves with sparsity rate $k$. RSI's recovery rate never reached 0.5.

\subsection{Creating our environment in Unreal Engine}

We test NATS in a pseudo-realistic environment using the Unreal Engine 4 (UE4) game development platform \cite{karis2013real} with the Airsim plugin \cite{shah2018airsim}. The UE4 platform allows the construction of different terrains and environments as well as the placement of objects within the environment. The Airsim plugin provides a Python API that allows the traversal of a vehicle through the UE4 environment. Airsim also allows collection of in-game data, such as first-person perspective screenshots of the environment and depth maps, two commonly available datasets in robotics applications \cite{engel2014lsd,mur2015orb,davison2007monoslam,meng2015ros}. Depth maps illustrate the distance between the camera and all objects in the environment. In practice depth maps could be obtained through different instruments and techniques the modeling of which is beyond the scope of this study (e.g. LIDAR or sonar tracking). 
Our UE4 environment consists of an approximately $500 \times 500$m treeless field with gentle hills that span an elevation range of $21$m. The field is surrounded by a forested, mountainous landscape.  We place models of different people within the environment along with models of various animals, including foxes, pigs, deer, wolves, and crows. The environment and all art assets and models were downloaded for free from the Unreal Marketplace.

\subsection{Mathematical Modelling of YOLOv3 Object Detector}
\begin{figure}
	\vspace*{1.5mm}
	\centering
	\begin{subfigure}{0.58\linewidth}
		\vspace*{2mm}
		\includegraphics[width=0.9\linewidth]{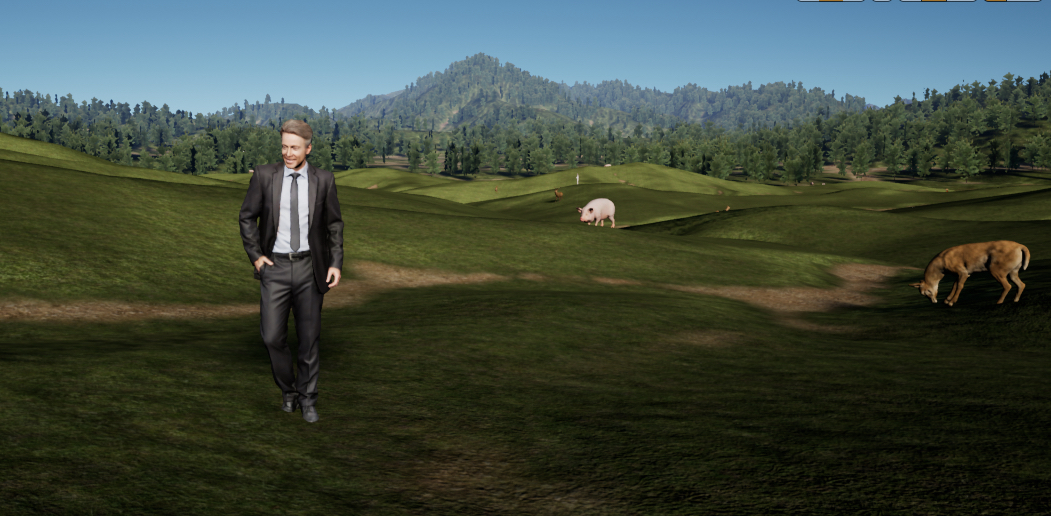}
		\vspace*{4mm}
		\caption{An example of a person in our\\ UE4 environment from the point \\of view of a robot}
		\label{fig:dennis}
	\end{subfigure}
	\begin{subfigure}{0.4\linewidth}
		\includegraphics[width=\linewidth]{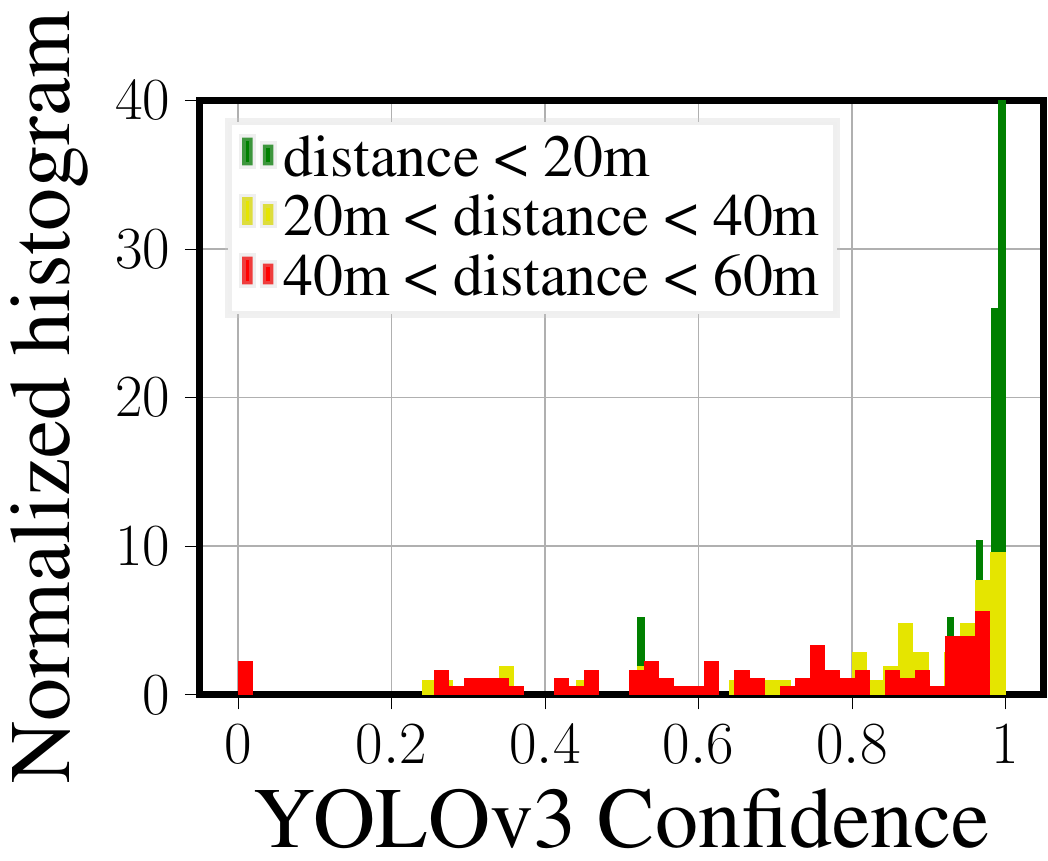}
		\caption{histogram on confidence score of YOLOv3 for 3 distances}
		\label{fig:hist}
	\end{subfigure}
	\caption{Object detection has been trained on COCO datasets and applied to our UE4 environment.}
\end{figure}

To back up Claim 1 in \fref{sec:formulation}, we randomly placed a large number of people and animals in our UE4 environment. Using AirSim, we generated about 100 image and depth maps from the created environment and checked the confidence score of YOLOv3 \cite{redmon2018yolov3} using the original weights trained by COCO dataset \cite{lin2014microsoft}. \fref{fig:dennis} shows an example of an image from our environment. Using this dataset, we created a normalized histogram as shown in \fref{fig:hist} of YOLOv3's confidence score on detected objects given their distance from the camera. \fref{fig:hist} clearly supports our mathematical modeling in \fref{fig:AccDistModel}.
Note that we could similarly write Claim 1 as how confidence changes with object's pixel size in an image as in \cite{dames2017detecting}. 
Using the pixel size can additionally allow modeling different camera focal lengths.
However, computing the number of pixels of an object requires segmentation methods with high accuracy which have high computational requirements \cite{jiao2019survey}. Instead we use fast online object detectors with bounding boxes. Our dataset is available online \cite{demo}.

\subsection{Apply NATS to our Unreal Engine Environment}
\begin{figure}
	\centering
	\includegraphics[width=1\linewidth]{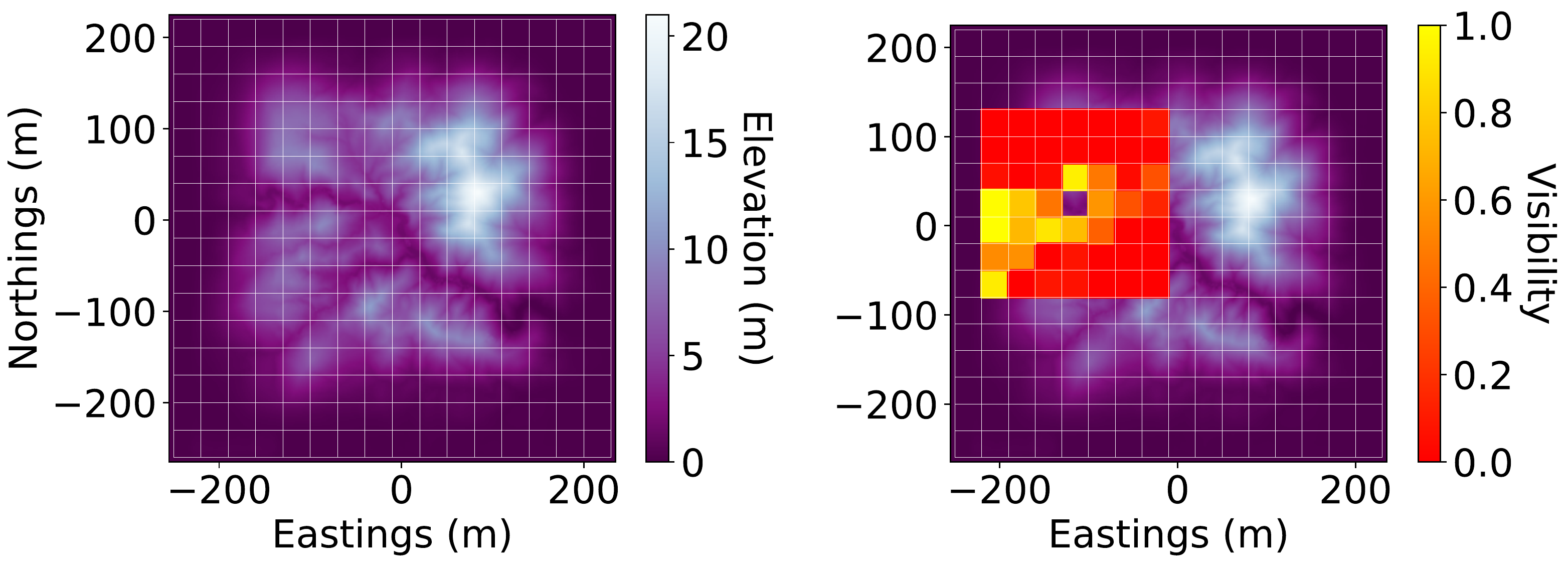}
	\caption{ (left) Topography of our UE4 environment in color with the coarse 30x30m grid overlain in white. (right) The percentage of each coarse grid that is visible to an agent located at -115 East, 25 North is shown as an example.}
	\label{fig:heightmap}
\end{figure}

We now test NATS's performance in our UE4 environment.
Since the environment is mountainous, sensing actions performed by ground robots can be partially obstructed from view by the hilly topography. 
We convert our UE4 environment to a geocoded Digital Elevation Map (DEM) with $1$m horizontal resolution (\fref{fig:heightmap}). We then create a coarse resolution coordinate system of the DEM using grid nodes spaced $30$m apart. The visible portions of the environment  (i.e., the viewshed) for a $2$m tall observer is calculated for all observation points in the coarse grid using the Geospatial Data Abstraction Library \cite{warmerdam2008gdal}.

We have included a video demonstration of NATS applied to our UE4 environment \cite{demo}. We have placed 6 different people randomly in the entire environment ($500\times500$m) to be found by 2 agents. Here, NATS considers travelling cost with $\alpha\!=\!1$. NATS successfully locates 5 out of 6 people at their correct location. Our video clearly demonstrates NATS's capability in getting closer to objects with lower confidence score. During the operating time in the video, a number of false positives appeared and were later refuted as the agents investigated them further. Some of the false positives that remain at the end of the simulation may be similarly temporary and could be refuted given a longer run time. Additionally, while YOLOv3 with original training generally performs well in our simulations, it is trained using images of real people (COCO) and not the simulacra used in our UE4 environment.  Performance can likely be improved by using a dataset explicitly trained on the images in our simulation.

We perform an experiment with 10 trials of randomly placing one person in a $250\times250$m area within our UE4 environment. We then compute the average distance travelled by each of two agents to locate the person. We compare the performance of two algorithms under this setting: 1) NATS considering YOLOv3’s uncertainty with distance using the confidence variance in Figure 5b, and 2) NATS disregarding this uncertainty by setting the confidence variance to 0. In both cases, the travelling cost in the reward is considered with $\alpha=1$. When NATS considers object detector uncertainty, each agent travels an average of 416m with standard error (SE) of 49m to find the person. Disregarding this information leads to a higher average of 563m with SE of 98m. Note that an exhaustive search would travel on average 1062m.


%
\section{Conclusions}
We have developed a new algorithm (NATS) for conducting active search using multiple agents that takes into account field sensor uncertainties. NATS does not need to know the number of objects of interest, it takes into account topography obstruction as well as travelling cost and manages communications between agents in a decentralized way. 
NATS performance improves accordingly with its number of agents and its complexity is not affected by either number of agents or targets. 
Future work includes considering moving targets which is useful for applications such as stopping animal poaching.
Finally, as part of an ongoing work, we intend to implement NATS on the real multi-robot search team in \cite{NREC}.


{
\bibliography{confs-jrnls,publishers,SearchAndMap}
}

\end{document}